\documentclass[11pt,a4paper]{article}
\usepackage{authblk}
\usepackage[hyperref]{eacl2021}
\usepackage{times}
\usepackage{latexsym}
\usepackage{xcolor}
\usepackage{textcomp}
\usepackage{graphicx}
\usepackage{hologo}
\usepackage{parskip}
\usepackage{array}
\usepackage{multirow}
\newcolumntype{C}[1]{>{\centering\arraybackslash}m{#1}}

\usepackage{microtype}
\usepackage{authblk}
\aclfinalcopy 

\title{L3CubeMahaSent: A Marathi Tweet-based Sentiment Analysis Dataset\\

}

\author{Atharva Kulkarni$^{1,3}$, Meet Mandhane$^{1,3}$, Manali Likhitkar$^{1,3}$, Gayatri Kshirsagar$^{1,3}$, \\\and 
Raviraj Joshi$^{2,3}$ \\

$^{1}$Pune Institute of Computer Technology, Pune\\
$^{2}$Indian Institute of Technology Madras, Chennai\\
$^{3}$ L3Cube, Pune\\

\texttt{\{k.atharva4899,meetmandhanemnm,manalil1806,gayatrimohan7\}@gmail.com}\\ 
\texttt{ravirajoshi@gmail.com}
  }

\date{}

\begin{document}
\maketitle
\begin{abstract}
Sentiment analysis is one of the most fundamental tasks in Natural Language Processing. Popular languages like English, Arabic, Russian, Mandarin, and also Indian languages such as Hindi, Bengali, Tamil have seen a significant amount of work in this area. However, the Marathi language which is the third most popular language in India still lags behind due to the absence of proper datasets. In this paper, we present the first major publicly available Marathi Sentiment Analysis Dataset - L3CubeMahaSent. It is curated using tweets extracted from various Maharashtrian personalities' Twitter accounts. Our dataset consists of  {\raise.17ex\hbox{$\scriptstyle\sim$}}16,000 distinct tweets classified in three broad classes viz. positive, negative, and neutral. We also present the guidelines using which we annotated the tweets. Finally, we present the statistics of our dataset and baseline classification results using CNN, LSTM, ULMFiT, and BERT-based deep learning models.
\end{abstract}

\textbf{Keywords:} Marathi Sentiment Analysis, Twitter Sentiment Dataset, Marathi Text Classification, Deep Learning, CNN, LSTM, BERT

\section{Introduction}

The use of social media has seen a sharp upward trend in recent years. It plays a big role in forming and shaping the views of people on various issues. From sharing facts and opinions to voicing dissent and grievances, the platform has gained popularity amongst many users \cite{nielsen2014relative}. Twitter is a significant social media platform. It has been quite popular in India for the past few years. It has been used by many politicians, journalists, and activists to connect with people directly. These kinds of interactions are generally strong on emotions, and can be used for developing sentiment analysis systems \cite{pak2010twitter,mathew2019spread}. Such systems have proven to be important for political analysis as well as identifying and curbing more complex issues such as fake news, harassment, hate speech, and bullying \cite{schmidt2017survey,joshi2021evaluation,wani2021evaluating}. In this work, we consider basic sentiment analysis or polarity identification tasks.

Popular languages such as English, Arabic, Russian, Mandarin \cite{rogers2018rusentiment,nabil2015astd,yu2020ch} as well as Indian languages such as Hindi, Bengali and Tamil have been explored on the sentiment task for a long time \cite{arora2013sentiment,patra2015shared,akhtar2016hybrid,mukku2017actsa,ravishankar2017corpus}. Many resources such as properly annotated datasets, SentiWordNets, annotation guidelines have been developed for these languages \cite{socher2013recursive,saif2013evaluation,hu2004mining}. Alternatively due to the low resource nature of many languages, translated versions of the English datasets were used for analysis \cite{joshi2019deep,refaee2015benchmarking,mohammad2016translation}. However, such translated datasets are often noisy due to the limitation of current translation systems for low resource languages.

Marathi is an Indian language spoken by around 83 million people and ranks as the third most spoken language in India. But surprisingly, there is no significant work or resource for the task of sentiment analysis in Marathi \cite{kulkarni2021experimental}. A sentiment analysis dataset curated by IIT-Bombay is available, but it has a very small size consisting of only 150 samples \cite{balamurali2012cross}. In this paper, we present L3CubeMahaSent\footnote{https://github.com/l3cube-pune/MarathiNLP} - the largest publicly available Marathi Sentiment Analysis dataset to date. This dataset is gathered using Twitter. The BERT based model fine-tuned on this dataset is also shared publicly\footnote{https://huggingface.co/l3cube-pune/MarathiSentiment}. Our work is summarized as follows:
\begin{enumerate}
    \item We present a {\raise.17ex\hbox{$\scriptstyle\sim$}}16,000 tweets strong Marathi Sentiment Analysis Dataset, manually tagged into three classes viz. positive, negative, and neutral.
    \item We provide a comprehensive annotation policy useful for tagging sentences by their sentiment. We also provide statistics for our dataset and a balanced split for experimentation.
    \item We present the result of our experiments on this dataset on recent deep learning approaches to create a benchmark for future comparisons.
\end{enumerate}

\section{Related Work}

Sentiment analysis is a fundamental task of Natural Language Processing \cite{medhat2014sentiment}. The absence of a proper sentiment analysis dataset for the Marathi language has led to limited research in this area. In this section, we will review some of the works introducing data resources in Indian and other languages. \newcite{balamurali2012cross} presented an approach for cross-lingual sentiment analysis using linked wordnets for Marathi and Hindi languages. For this purpose, they used various blogs and travel editorials as a dataset which consisted of about 75 positive and 75 negative reviews. The WordNet approach showed an improvement of 14-15 percent over the approach using a bilingual dictionary. The Marathi dataset created in this work is very small and cannot be used to train existing deep learning algorithms.


\newcite{bhardwaj2020hostility} presented a hostility detection dataset in Hindi. Data was collected from various online platforms like Twitter, Facebook, Whatsapp, etc., and was benchmarked using machine learning algorithms namely, support vector machine (SVM), decision tree, random forest, and logistic regression. They also labeled each hostile post as either fake, hateful, offensive, or defamation.

\newcite{patra2018sentiment} presented details of a shared task in a competition on sentiment analysis of code-mixed data pairs of Hindi-English and Bengali-English. The best performing team used SVM for sentence classification. The sentiment analysis of code-mixed English-Hindi and English-Marathi text is also studied in \cite{ansari2018sentiment}.

\newcite{nabil2015astd} introduced Arabic sentiments tweets dataset consisting of 10,000 tweets classified as objective, subjective positive, subjective negative, and subjective mixed. They tried 4-class sentiment analysis as well as 3-class sentiment analysis on the dataset and found that the former was more challenging. They also concluded that SVM performed well on the dataset for the task.

\newcite{rogers2018rusentiment} presented RuSentiment, a dataset for sentiment analysis in the Russian language. They performed experiments on their dataset using algorithms like logistic regression, linear SVM, and neural networks. The best performance was observed in the case of neural networks. They also released the fastText embeddings that they have used for experimentation.

\newcite{ikoro2018analyzing} presented results of analyzing sentiments of UK energy consumers on Twitter. They proposed a method in which they combined functions from two sentiment lexica. The first lexicon was used to extract the sentiment-bearing terms and the negative sentiments. The second lexicon was used to classify the rest of the data. This method improved the accuracy compared to the general method of using one lexicon. 

\section{Curation of Dataset}

\subsection{Dataset Collection}

For creating our dataset, we first manually created a list of various famous personalities who actively tweet about current affairs. Twitter profiles were shortlisted based on their frequency, relevance of activity, and degree of the sentiment of the tweets. Hence a majority of the tweets are from political personalities' profiles and activists as they express a wide range of emotions and sentiments. We attempted to improve the diversity of the points of view contained in the dataset.

All tweets in this dataset are specifically in the Marathi language. All hashtags, mentions, special symbols, and the occasional English words are kept in the tweets in the publicly available version of the dataset. We think it is best to keep the original dataset unhampered for anyone to experiment on it. However, while performing experiments, we have removed the aforementioned tokens from the tweets during data pre-processing. Also, the dataset does not retain any context of the tweets such as the tweeting profile, time of posting, region, etc.

As far as scraping the tweets is concerned, there are multiple python libraries available. Some of them are Tweepy (the official open-source library provided by Twitter)\footnote{https://www.tweepy.org/}, GetOldTweets3\footnote{https://pypi.org/project/GetOldTweets3/}, Twint\footnote{https://pypi.org/project/twint/}, and Snscrape\footnote{https://github.com/JustAnotherArchivist/snscrape}. We have used the Twint library to scrape tweets.

\subsection{Dataset Annotation}

We have manually labeled the entire dataset into three classes: positive, negative, and neutral. These three classes have been denoted by `1',`-1', and `0' respectively. The dataset was split among the entire team to tag in parallel. In order to maintain consistency while tagging tweets, we developed an annotation policy.

To begin with, we ensured not to take into account the author of the tweet, thereby eliminating any bias towards any author. Tweets are tagged by a general assumption that they are posted by any random person. Positive emotions such as happiness, gratitude, respect, inspiration, support are tagged as positive. Negative emotions such as hate, disrespect, grief, insult, disagreement, the opposition are tagged as negative. Tweets that do not convey a strong feeling, such as simple facts, statistics, or statements are tagged as neutral.

Tweets containing sarcasm, irony which clearly depict a negative sentiment are tagged as negative. Congratulatory and thank-you tweets are tagged as positive. A tweet that criticizes something or someone, or which states a fact stating an adverse event or reaction is termed negative. However, if the criticism comes as constructive and healthy, mentioning possible solutions, then it is tagged as positive. Finally, tweets containing mixed sentiments are labeled by the more dominant emotion expressed.

Even though these rules were laid down, there were some tweets that simply were difficult to tag by a single individual and needed to be reviewed. In such cases, we took a vote amongst the team and tagged the tweet according to the majority votes. Tweets for which no consensus could be formed were removed.

Some examples of tagged tweets are mentioned for more clarity in Table
3. 

\begin{table}
\begin{center}
\begin{tabular}{{p{1.5cm}p{1cm}p{1cm}p{1cm}p{1cm}}}
\hline
\textbf{Split}&\textbf{Total Tweets}&\textbf{1}&\textbf{-1}&\textbf{0} \\
\hline
Train& 12114 & 4038 & 4038 & 4038  \\
Test & 2250 & 750 & 750 & 750 \\
Validation & 1500 & 500 & 500 & 500 \\
\hline
\end{tabular}
\caption{\label{stat-tab} Dataset Statistics. }
\end{center}
\end{table}

\begin{figure}
\centerline{\includegraphics[width=1\columnwidth]{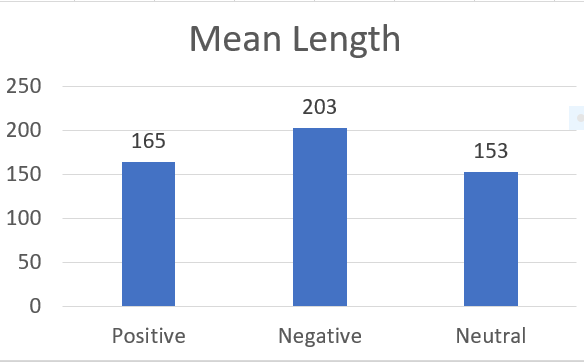}}
\caption{Mean length of records per class (in words)}
\label{avg_len}
\end{figure}

\subsection{Dataset Statistics}

\begin{figure*}
\centerline{\includegraphics[width=1.2\columnwidth]{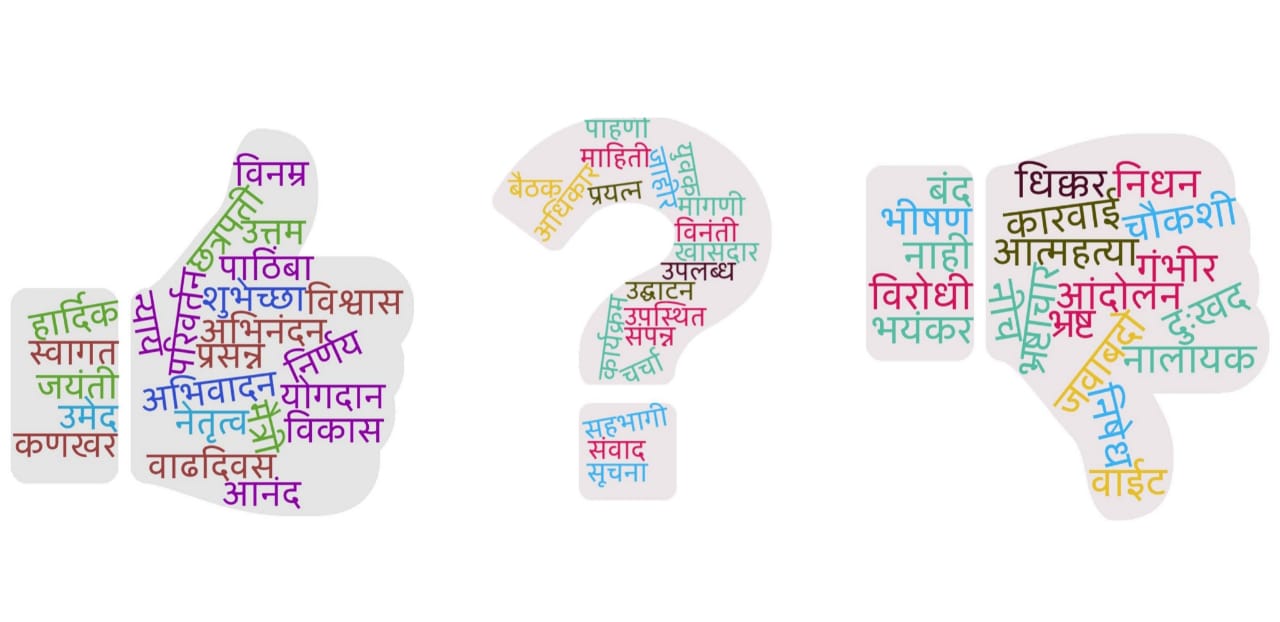}}
\caption{Positive, Neutral and Negative wordclouds}
\label{wc}
\end{figure*}

Initially, we annotated a total of 18,378 tweets. But, in order to ensure that the classes are balanced, we randomly selected an equal number of tweets for each class. Hence, the final version of L3CubeMahaSent consists of 15,864 tweets. Table \ref{stat-tab} shows class-wise distribution and the train-test-validation split. The remaining 2,514 annotated tweets will also be published along with the dataset. It consists of 2,355 positive and 159 negative tweets. These extra tweets have not been used for model evaluation. Commonly occurring words in each class can be visualized in the form of word-clouds as shown in Figure \ref{wc}.

\section{Evaluation}

\subsection{Experimentations}
We performed 2-class and 3-class sentiment analysis on our dataset. For conducting baseline experiments on our dataset, hashtags, mentions, and special symbols were removed during preprocessing. We used some of the widely used text classification architectures for sentiment classification \cite{kulkarni2021experimental,kowsari2019text,kim2014convolutional,sun2019fine}. The text is tokenized as words or sub-words and passed to the algorithms mentioned below:

\begin{itemize}
    \item \textbf{CNN: }The initial embedding layer outputs word embeddings of size 300. These embeddings are passed to a Conv1D layer having 300 filters and kernel size 3. A global max-pooling is applied to the output sequences to get a sentence representation. This is then passed on to a dense layer having size 100. A final dense layer having size equal to the number of classes is added to give classification results. We have experimented with various types of embedding layers having random initialization (word and subword), original Facebook fastText embeddings (trainable and non-trainable) \cite{mikolov2017advances}, and Indic fastText embeddings (trainable and non-trainable) by IndicNLP \cite{kakwani2020inlpsuite}.
    
    \item \textbf{BiLSTM+GlobalMaxPool: }This is similar to the CNN network with Conv1D layer replaced by a Bi-LSTM layer. Inputs are fed to an embedding layer which outputs word embeddings of size 300. These embeddings are given to a bi-directional LSTM layer with cell size 300 and then output is max pooled over time. A dense layer of size 100 and a subsequent dense layer of size equal to the number of classes complete the architecture. Embeddings same as those mentioned in the CNN section are also experimented with.
    
    \begin{table*}
\begin{center}
\begin{tabular}{{p{2cm}p{5cm}p{2cm}p{2cm}}}
\hline \textbf{Model} & \textbf{Variant} & \textbf{3-class Accuracy} & \textbf{2-class Accuracy} \\
\hline
\multirow{6}{*}{CNN} & random-word & 79.47 & 90.00\\
& random-subword &  81.56 & 91.73\\
& FB fastText-Trainable & 81.02 & 92.67\\
& FB fastText-Static &  80.18 & 90.93\\
& Indic fastText-Trainable & \textbf{83.24} & \textbf{93.13}\\
& Indic fastText-Static & 83.00 & 92.47\\
\hline
\multirow{6}{*}{BiLSTM} & random-word & 80.93 & 90.87 \\
& random-subword & 79.42 & 89.80\\
& FB fastText-Trainable & 81.78  & 92.33\\
& FB fastText-Static & 79.87  & 89.67\\
& Indic fastText-Trainable & \textbf{82.89} & 91.8\\
& Indic fastText-Static & 82.41 & \textbf{92.67}\\
\hline
ULMFiT & (iNLTK) & 80.80 & 91.40\\
\hline
\multirow{2}{*}{BERT} & mBERT & 80.66 & 91.40\\
& IndicBERT (INLP) & \textbf{84.13} & \textbf{92.93}\\
\hline
\end{tabular}
\caption{\label{result-tab} Classification accuracies over different architectures. The 2-class accuracy corresponds to positive and negative class only.}
\end{center}
\end{table*}

    \item \textbf{ULMFiT: }ULMFiT is also a LSTM based model \cite{howard2018universal}. It uses transfer learning which allows the model to be finetuned quickly on the target dataset using even a small sample set. We use a publicly available ULMFiT model for the Marathi language released by iNLTK and finetune it on our dataset \cite{arora2020inltk}.
    
    \item \textbf{BERT: }The BERT is a transformer based model pretrained on a huge text corpora, which can be finetuned for any target dataset \cite{devlin2018bert}. Many publicly available flavours of BERT are available, and we use two specific multilingual models:
    \begin{itemize}
        \item Multilingual-BERT (mBERT)
        \item Indic-BERT by IndicNLP \cite{kakwani2020inlpsuite}
    \end{itemize}
    For both of these models, we used the CLS token for sequence classification.
\end{itemize}

\subsection{Results}

We experimented with a variety of architectures such as CNN and BiLSTM for text classification on our dataset along with different embeddings. We have used random word and subword initializations, and also used pre-trained word embeddings made public by Facebook and IndicNLP. Both of these pre-trained embeddings were used in trainable and static modes.
Along with these architectures, pre-trained models such as ULMFiT, mBERT, and IndicBERT were also used.

The results from our experiments were in line with previous works done in Marathi text classification. Pretrained word embeddings give a definite edge over random initializations. The CNN-based models have a slight advantage over BiLSTM based models. It was observed that though models using random initializations generally give good results, they tend to quickly overfit while training. The use of pre-trained embeddings significantly reduces overfitting. The results for 3-way and 2-way classification are shown in Table \ref{result-tab}. The neutral class is dropped for 2-way classification.

\begin{figure}
\centerline{\includegraphics[scale=0.75]{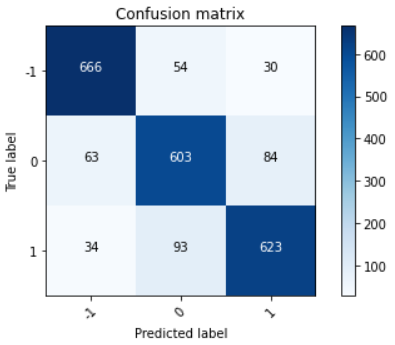}}
\caption{Confusion Matrix for IndicBERT - 3 class classification}
\label{conf-mat-3way}
\end{figure}

\begin{figure}
\centerline{\includegraphics[scale=0.75]{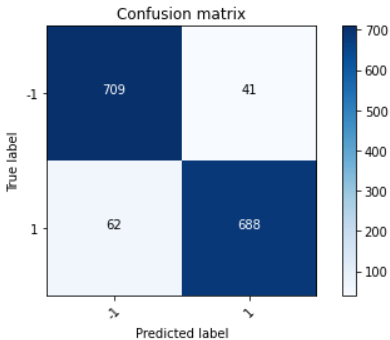}}
\caption{Confusion Matrix for CNN with Indic fastText (trainable) - 2 class classification}
\label{conf-mat-2way}
\end{figure}

The Marathi word embeddings provided by IndicNLP perform better than the original versions released by Facebook. Keeping word embeddings trainable further increases the accuracy. The ULMFiT gives results that are comparable with simple CNN and BiLSTM models. The CNN model combined with trainable Indic fastText word embeddings gives the best results in the 2-class classification experiments, slightly outperforming the IndicBERT. The IndicBERT was the best performing model for the more difficult 3-class classification experiments. The respective confusion matrices are shown in Figure \ref{conf-mat-2way} and Figure \ref{conf-mat-3way}.

\begin{figure*}
\centerline{\includegraphics{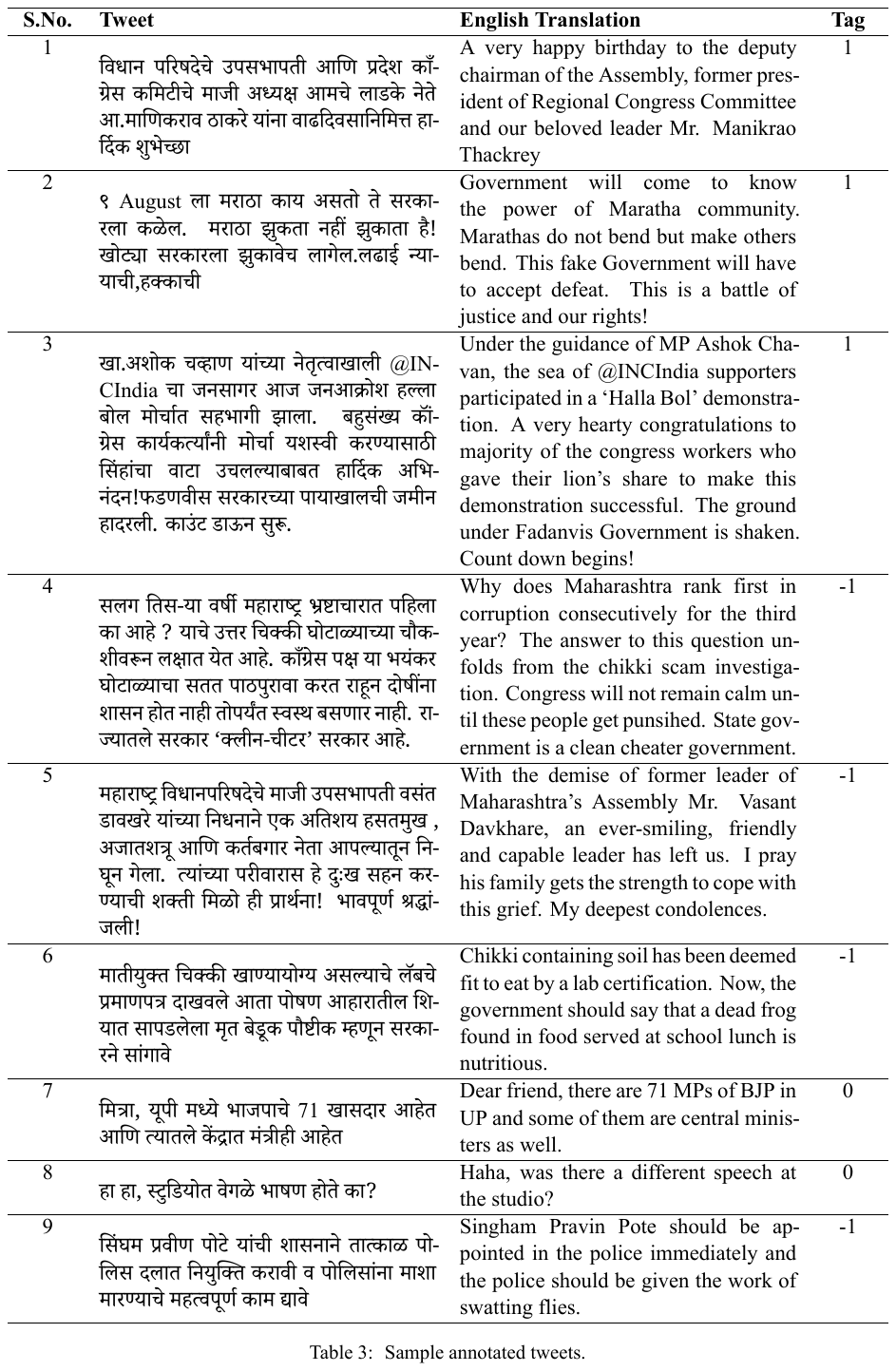}}
\end{figure*}

\section{Conclusion}

In this paper, we have presented L3CubeMahaSent - the first major publicly available dataset for Marathi Sentiment Analysis which consists of {\raise.17ex\hbox{$\scriptstyle\sim$}}16000 distinct tweets. We also describe the annotation policy which we used for manually labeling the entire dataset. We performed 2-class and 3-class sentiment classification to provide a benchmark for future studies. The deep learning models used for sentiment prediction were CNN, Bi-LSTM, ULMFiT, mBERT, and IndicBERT. The publicly available Marathi fastText embeddings were used with word-based models. We report the best accuracy using IndicBERT and CNN with Indic fastText word embeddings. We hope that our dataset will play a crucial role in advancing NLP research for the Marathi language.

\section*{Acknowledgments}

This work was done under the L3Cube Pune mentorship program. We would like to express our gratitude towards our mentors at L3Cube for their continuous support and encouragement.

\bibliography{anthology,eacl2021}
\bibliographystyle{acl_natbib}

\end{document}